\documentclass[journal]{IEEEtran}
\pdfoutput=1
\ifCLASSINFOpdf
\else
\fi

\usepackage{url}            % simple URL typesetting
\usepackage{booktabs}       % professional-quality tables
\usepackage{amsfonts}       % blackboard math symbols
\usepackage{nicefrac}       % compact symbols for 1/2, etc.
\usepackage{microtype}      % microtypography
\usepackage[space]{grffile} % for using space and underscore in filename
\usepackage{subcaption}
\usepackage{graphicx}
\usepackage{amsmath}
\usepackage{cite}

\begin{document}
\title{Applying Topological Persistence in Convolutional Neural Network for Music Audio Signals}


%\author{Michael~Shell,~\IEEEmembership{Member,~IEEE,}
%        John~Doe,~\IEEEmembership{Fellow,~OSA,}
%        and~Jane~Doe,~\IEEEmembership{Life~Fellow,~IEEE}% <-this % stops a space
%\thanks{M. Shell was with the Department
%of Electrical and Computer Engineering, Georgia Institute of Technology, Atlanta,

\author{
  Jen-Yu~Liu, 
  Shyh-Kang~Jeng, 
  Yi-Hsuan~Yang
\thanks{J.-Y. Liu (contact: ciauaishere@gmail.com) and S.-K. Jeng are with Graduate Institute of Electrical Engineering, National Taiwan University, Taiwan}%
\thanks{J.-Y. Liu and Y.-H. Yang are with CITI, Academia Sinica, Taiwan}}

\maketitle


\begin{abstract}
Recent years have witnessed an increased interest in the application of persistent homology, a topological tool for data analysis, to machine learning problems. Persistent homology is known for its ability to numerically characterize the shapes of spaces induced by features or functions.
On the other hand, deep neural networks have been shown effective in various tasks. To our best knowledge, however, existing neural network models seldom exploit shape information. In this paper, we investigate a way to use persistent homology in the framework of deep neural networks. Specifically, we propose to embed the so-called ``persistence landscape,’’ a rather new topological summary for data, into a convolutional neural network (CNN) for dealing with audio signals.
Our evaluation on automatic music tagging, a multi-label classification task, shows that the resulting persistent convolutional neural network (PCNN) model can perform significantly better than state-of-the-art models in prediction accuracy. We also discuss the intuition behind the design of the proposed model, and offer insights into the features that it learns.

\end{abstract}


\section{Introduction}
People have adopted different strategies to extract features in neural networks. Recurrent neural networks (RNNs) consider the information from the past or/and the future of a signal by using recurrent connections, which have yielded huge success in speech recognition \cite{graves12book}. On the other hand, convolutional neural networks (CNNs) include context information by taking weighted sum in a segment of a given length \cite{krizhevsky12imagenet}. In this paper, we intend to include the shape information of a segment into the neural network, by using a topological tool called persistent homology. %into CNN for music audios.

The goal of persistent homology is to describe the shape (e.g. connected components or holes) of a set of data points \cite{edelsbrunner08ph}. To achieve this, it forms a nested sequence of sub-objects such that the latter ones includes the former ones and each sub-object is described topologically. The end product from persistent homology is usually referred to as a \emph{summary} for the data shape.
The most commonly used summaries include \emph{persistence diagram} \cite{edelsbrunner02}, \emph{barcode} \cite{carlsson04}, and the recently proposed \emph{persistence landscape} \cite{bubenik15pl, bubenik13toolbox}. A nice property of the summaries is that they are stable with respect to perturbations of the data \cite{edelsbrunner08ph}. Such summaries can be employed to discern the topological difference between two sets of data. For example, a recent work done by Reininghaus \emph{et al.} establishes a multi-scale kernel for persistence diagram that can be used with kernel-based classification algorithms such as support vector machine for 3D shape classification \cite{reininghaus15kernel}.

The use of persistence homology has been found beneficial for a number of machine learning tasks in recent years. For example, Li \emph{et al.} \cite{li14cvpr} also used persistence diagram for 3D shape recognition.
 
In the audio domain, Brown \emph{et al.} \cite{brown09} used persistence barcode on the raw signals of speech to learn structural features that distinguish among vowels, nasals, and noisy sounds. Bergomi \cite{bergomi15} applied persistence to a tone representation % called Tonnetz
to analyze the structure of music pieces. % and clustered the data accordingly. 
Emrani \emph{et al.} \cite{emrani14} used barcode for wheeze detection from breathing sound signals.

Although the persistence homology has shown potential in various areas, it has not been incorporated in a neural network model, to the best of our knowledge. 

Motivated by the potential gain of combining the idea of persistence homology and deep learning, we pioneer this research front by firstly designing a neural network model to characterize the topological features of audio signals. Specifically, we propose a design that exploits persistence landscape in a CNN.

We choose to work with CNN for its well demonstrated discriminative power in various audio-related tasks such as speech recognition \cite{abdel14cnnspeech} and music structure analysis \cite{ullrich14boundary}. 
In addition, as will be elaborated in Section \ref{sec:model}, it is easy to combine the outputs of a convolutional layer and a dedicated layer for persistence homology in the segment level, an intermediate level for audio signal processing.

We evaluate the proposed persistent convolutional neural network (PCNN) model on the task of music auto-tagging, a multi-label classification task that aims at assigning tags such as genres and instruments to music pieces \cite{turnbull08tagging, tingle10mm}.

The state-of-the-art method for music auto-tagging proposed by 
Dieleman \emph{et al.} \cite{dieleman13multiscale} uses a convolution-flavored feature processed by principal component analysis and clustering. Furthermore, they proposed different ways to exploit multi-scale information. 
As an extension of this method, Dieleman \emph{et al.} later applied CNN on the raw audio signals and log mel-spectrograms and achieved similar accuracy for music auto-tagging. Our evaluation shows that PCNN can outperform these two methods.

The remainder of the paper is organized as follows. Section \ref{sec:pl} introduces persistence landscape. We present the proposed model in Section \ref{sec:model}. Then, we show the experimental results for music auto-tagging and our observations in Section \ref{sec:exp}. Finally, we conclude this paper in Section \ref{sec:conclusion}.

\section{Persistence Landscape} \label{sec:pl}

The topological summary used in this paper is persistence landscape proposed by Bubenik \cite{bubenik15pl, bubenik13toolbox}. In persistent homology, the target object is often topologically represented in different ways, depending on the target applications. In this paper, we consider \emph{cubical complexes} for audio signals are basically constructed on equal-spaced grids \cite{wagner12cubical, kaczynski04comphomo}. Specifically, we use one-dimensional cubical complexes, $\mathbb{T}=\{0, 1, 2, ..., N\}$ and the edges connecting two neighbors, denoted as $\{(i, i+1)\}_i$, where $N$ is the length of a feature sequence for a audio segment. One of the benefit of cubical complex is that it is naturally equipped with the notion of \emph{connectivity}. We consider 2-connectivity in this paper, where each element $t$ in $\mathbb{T}$ connects to its immediate neighbors $t-1$ and $t+1$. Due to space limit, we will not formally introduce persistent homology but instead provide the intuitions behind the definitions. For an introduction to homology and persistent homology, please refer to \cite{edelsbrunner08ph,kaczynski04comphomo}.

In homology theory, homology classes are used to characterize non-boundary cycles. Persistent homology studies the change of homology classes constructed from $\mathbb{T}$ and a \emph{filtering function} $f: \mathbb{T} \rightarrow \mathbb{R}$. In the case considered here, the values of function $f$ are provided by the output signals from a convolution layer. We can construct from $f$ a chain of sub-complexes that starts with the empty complex and ends with the complete complex, i.e., $\emptyset=\mathbb{T}_{v_0} \subset \mathbb{T}_{v_1} \subset \dots \subset \mathbb{T}_{v_m} = \mathbb{T}$, where $v_i$s are real values. 

Each sub-complex $\mathbb{T}_{v_i}$ is defined as the superlevel-set $\mathbb{T}_{v_i}:=f^{-1}([v_i, \infty))$ \cite{edelsbrunner08ph, li14cvpr}. In the processing of audio features, we care more about the higher values of the features because the parts of higher values often means there are some interesting events. We only consider 1-dimensional cubical complex, so the only possible non-trivial homology classes are 0-dimensional homology classes, which describe connected components.

\begin{figure*}[!t]
\centering
\includegraphics[width=1.0\textwidth]{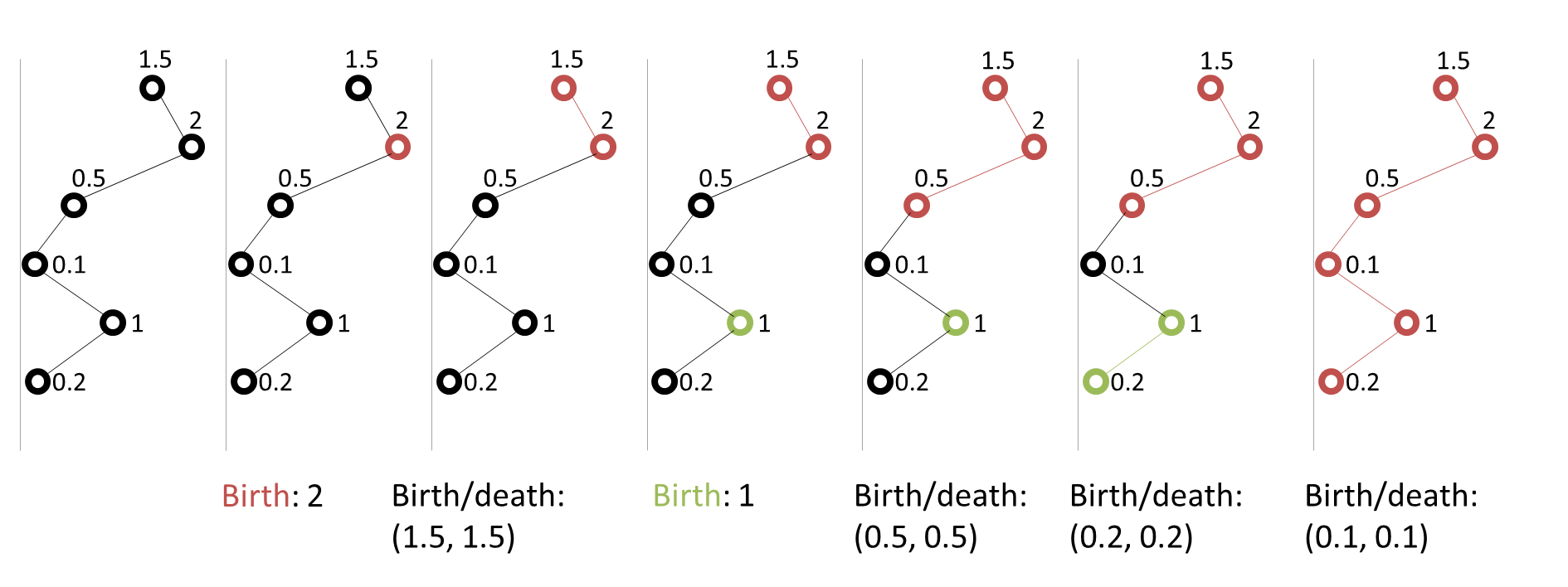}
\caption{Derivation of births and deaths. The process goes from left to right. There are two more birth-death pairs formed in the right-most stage: $(1, 0.1)$ for the green component and $(2, -\infty)$.}
\label{fig:bd}
\end{figure*}

The next step is to derive the \emph{births} and \emph{deaths} of homology classes, which are, in our case, the births and deaths of connected components. A birth is a value where a new component appears, and a death is a value where a component is merged into an earlier born component, as illustrated in Figure \ref{fig:bd}. 

The \emph{persistence} is the difference between the death and birth times. Therefore, the birth-death pairs are the basis of either barcode, persistence diagram, and persistence landscape.

\begin{figure*}[!t]
\centering
\includegraphics[width=\textwidth]{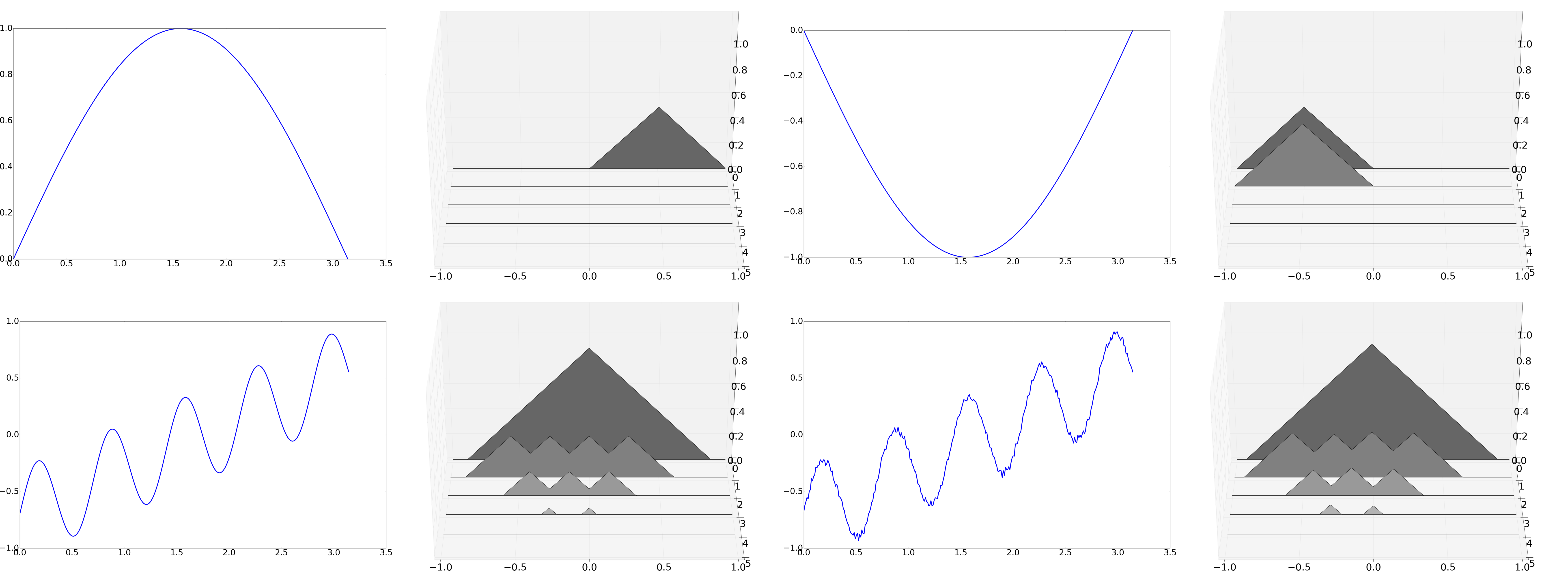}
\caption{Various Signals signals and the corresponding persistence landscapes.}
\label{fig:sigpl}
\end{figure*}

Given a birth-death pair $(b, d)$, we can construct the persistence landscape as follows \cite{bubenik13toolbox}. First, piece-wisely linear functions $f_{(b, d)}$ are constructed:
\begin{equation}
\label{eq:f}
f_{(b, d)}=
\begin{cases}
0 & \quad \text{if } x \not\in (d, b) \,,\\
x-d & \quad \text{if } x \in (d, \frac{d+b}{2}] \,,\\
-x+b & \quad \text{if } x \in (\frac{d+b}{2}, b) \,.\\
\end{cases}
\end{equation}
It can be seen that $f_{(b, d)}$ has a triangle shape. Given the birth-death pairs of a space, the persistence landscape is defined as the functions $\lambda_k:\mathbb{R}\rightarrow[0, \infty]$, where $\lambda_k(x)$ is the $k$-th largest value of the sequence $\{f_{(d_i, b_i)}(x)\}_{i=1}^n$. We will refer to $\lambda_k(x)$ as the $k$-th piece of a persistence landscape. The global maximum of $f$ on the space will not have a finite death time in theory. In practice, we assign the global minimum to the death of the component of the global maximum, as done in \cite{li14cvpr}. Some examples of signals and the corresponding persistence landscape are shown in Figure \ref{fig:sigpl}. For the signal on the top left, there is one local maximum so the persistence landscape is a single mountain. For the signal on the top right, there are two local maxima of the same value so there are two mountains. One benefit of persistence landscape is its invariance to small noises, as demonstrated from the two signals at the bottom of Figure \ref{fig:sigpl}.

\subsection{Intuition about Persistence Landscape}
We may consider the first component as providing an overview of the segment. In our setting, the first component of the persistence landscape always contains a mountain where the left foot is the global minimum and the right foot is the global maximum. We may think of the peaks in the persistence landscapes as emphasizing the more stable part of a component, sitting right between the birth and the death of a birth-death pair. In other words, it encodes the most prominent part of all the $f_{(b, d)}$.

If $\lambda_k$ in a persistence landscape is nonzero at some value $v$, it implies that there is a $\mathbb{T}_i$ that has at least $k$ connected components that containing $v$. For an audio signal, this means that there is at least $k$ local maxima. Therefore, a signal has a nonzero value in $\lambda_k$ with higher $k$ indicates larger fluctuation.

\subsection{Why Persistence Landscape instead of Persistence Diagram?}

Another popular persistence summary is the persistence diagram, which is defined as the set of tuples of birth-death pairs. The relationships between two persistence diagram can be measured through the bottleneck metric or p-Wasserstein metric \cite{li14cvpr, fasy14pd, kwitt15kernel}. We argue below why persistence landscape is more suitable for being used in a network based model as compared with persistence diagram.

In a network, it is convenient if we can have a representation of matrix form by which the pointwise comparison of two matrices is meaningful. However, persistence diagram does not have this property. In contrast, we can do pointwise addition on persistence landscapes because they are essentially functions \cite{bubenik15pl}. Moreover, as a persistence landscape is a function, for computational purposes we can convert these functions to matrices by subsampling the persistence landscape in a chosen range of the domain. We may think of this representation as a restriction of the persistence diagram functions to a subsets of the domain, so the addition can still be done pointwisely. In this way, we can simply treat the persistence landscape as a finite-size, two-dimensional feature map that can be easily processed by a subsequent convolutional layer in a CNN architecture. While the subsampling can be done in various ways, for simplicity a uniform sampling approach is adopted in this work.

\subsection{Persistence Landscape and CNN}
CNN provides a good environment for the incorporation of persistence landscape. In common CNNs, a convolution layer uses a number of filters to process an input signal and then summarizes the information by using max-pooling on segments of finite length $S$. The assumption here is that these summaries of the segments provide information useful to the subsequent processing. On the other hand, a landscape is computed over a topological space, a cubical complex in our case. With a \emph{persistence layer}, we can compute the persistence landscape of the same segments of length $S$ from the input signal, thereby offering a different way to characterize the content of the input signal.  
As the information captured by this persistence layer and a convolution layer might be complementary to each other, we can also concatenate the features derived from the two processing pipelines. This can be easily implemented as their outputs can have the same number of temporal units.

\section{Models} \label{sec:model}

\begin{figure}[!t]
\centering
\includegraphics[width=0.4\textwidth]{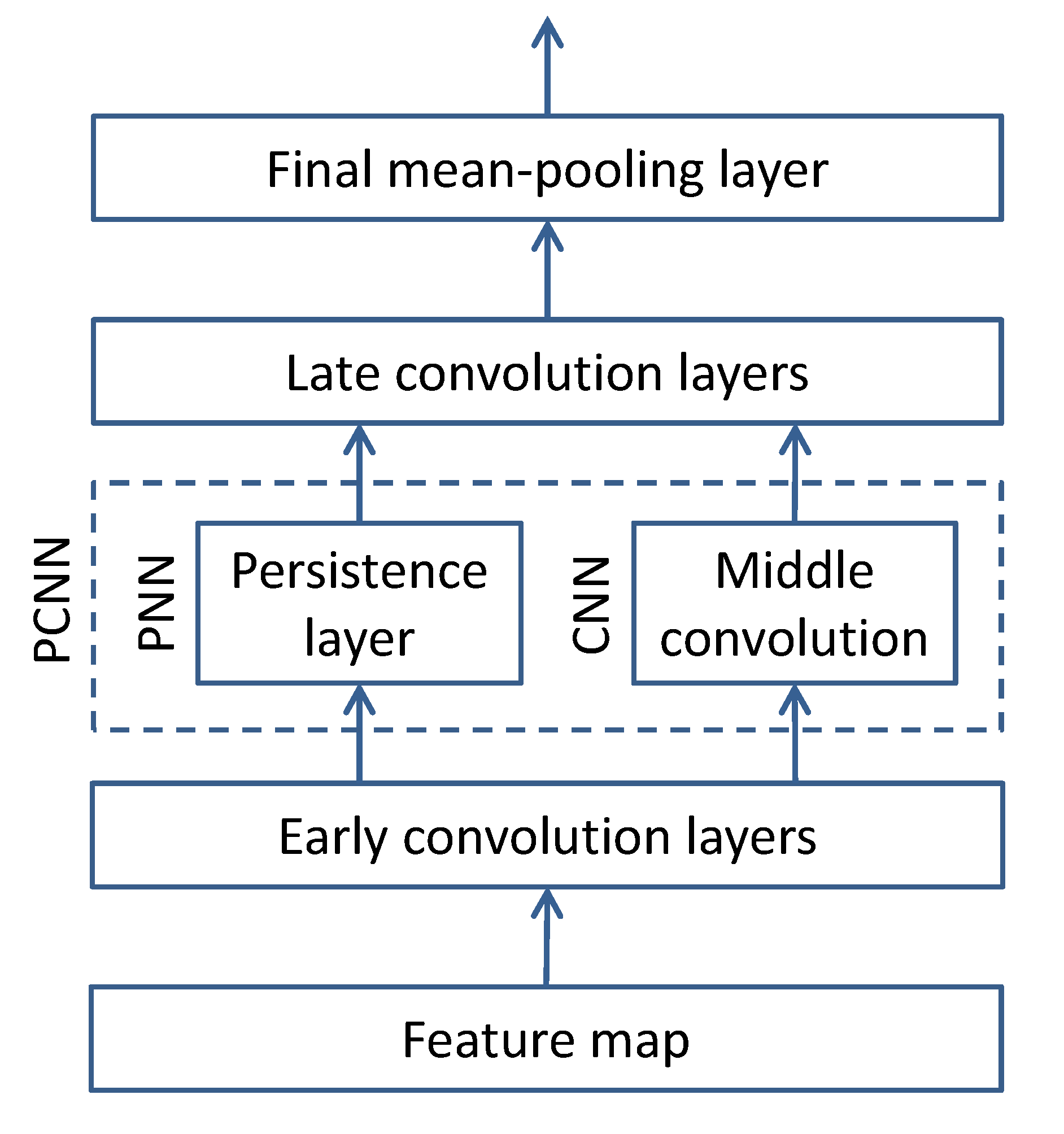}
\caption{Network structure}
\label{fig:structure}
\end{figure}

We show the network structure in Figure \ref{fig:structure}. We use multiple times of convolution layers in the structure. A $(K, L, S)$ convolution layer consists of a convolution of $K$ filters and convolving size $L$ and a max-pooling sub-layer that performs max pooling every $S$ units along time axis. The input to the network is a feature map, a matrix with a temporal axis and a feature axis. The feature map is first processed by a stack of convolution layers, referred to as the \emph{early convolution} layers.

The output of the final early convolution layer is fed into either a middle convolution layer, a persistence layer, or both. A middle convolution layer is a $(K_m, L_m, S_m)$ convolution layer. A network uses only middle convolution without persistence layer is simply a CNN. A persistence layer processes each filter from the preceding layer separately, each time using a segment of length $T$. We can view each filter as the filtering function $f$. A persistence layer will have the following fixed parameters, a \emph{value range} $(C_0, C_1)$ deciding the range to sample a persistence landscape, the \emph{number of pieces} $P$, deciding how many pieces in a persistence landscape we will use, and the \emph{number of sample points} $Q$, deciding how many points we should sample uniformly from the value range. We can therefore refer to a persistence layer as a $(C_0, C_1, P, Q)$ layer. If there are $U$ filters from the output of the early convolution layers, the total feature dimension will be $P\times Q\times U$. A network that uses only the persistence layer in this part is referred to as the persistent neural network (PNN). We can also combine the outputs of the middle convolution layer and the persistence layer by concatenation, leading to the persistent convolutional neural network (PCNN). For PCNN, we set $S_m=T$ to ensure that the temporal scales from the two layers are the same.

The output of from the previous layer is processed by another stack of convolutions, referred to as the \emph{late convolution} layers. The late convolution layers use $L=1$ and $S=1$ (i.e. no temporal context and no max poolings),

thereby providing the function of the fully connected layers in conventional CNNs. With this replacement, we can process sequences of arbitrary length \cite{pathak14fcn, oquab15localization}. The last late convolution layer is the \emph{segment output sub-layer}, where the number of units if equal to the number of tags. Its outputs are pooled temporally with the \emph{final mean-pooling} layer to give a final prediction for the entire music clip.

We put the persistence layer after a convolution layer for two reasons. First, the convolution layers can serve as a dimensionality reduction device as the persistence operation is more computationally expensive. Second, we want the information can back-propagated through persistence layer so that the persistence layers have influences on the learned features.

\subsection{How Back-propagation Works through the Persistence Layer}

Persistence landscapes are constructed from piece-wise linear functions $f_{(b_i, d_i)}$. The values of $f_{(b, d)}$ are composed of linear functions of $b$ and $d$ in a birth-death pairs $(b, d)$, as shown in Equation (\ref{eq:f}). A persistence landscape is simply a re-ordering of the function values in its sampled matrix form. Note that the deaths and births are all local extrema. For an element in a persistence landscape matrix, the back-propagation is done through the elements which own the birth or death value.

\section{Evaluation} \label{sec:exp}
We evaluate on a music auto-tagging dataset MagnaTagATune \cite{law09magna}. It contains multi-label annotations collected from human evaluations of music tags through playing a game. There are totally 188 tags and 25,863 29-second clips. It includes tags of instruments, tempo, genres, acoustic, etc. 
We note that there are at least two versions of MagnaTagATune and some people used an earlier 160-tag version \cite{hamel11, nam15}. 
However, this version is not publicly available. We instead use the current publicly available version and follow the setting employed in \cite{dieleman13multiscale, dieleman14icassp} --- while MagnaTagATune natively has 16 sub-folders with no overlapping artists, we use the 1st--12th sub-folders for model training, the 13th sub-folder for validating and parameter tuning, and the 14-16th sub-folders for testing. Following \cite{dieleman13multiscale, dieleman14icassp}, we consider the top 50 tags (out of the 188 possible tags) in most experiments.

Moreover, following the convention in the literature, we use the average area under the ROC-curve (AUC) and mean average precision (mean AP; or MAP) as the performance metrics \cite{turnbull08tagging, dieleman13multiscale}. The evaluation can either be done per-class or per-clip. The per-class result computes AUC or AP for each tag class across all the clips, whereas the per-clip result computes AUC or AP in a clip over all tags.

We extract 128-D log mel-spectrogram from the audio, which has been used extensively in recent work on music auto-tagging \cite{hamel11, dieleman13multiscale, dieleman14icassp}. This is to facilitate our comparison with existing work in terms of the learning models, rather than the input features. 
The features are z-score normalized using parameters 

estimated from the training set. The mel-spectrograms are extracted with the librosa library \cite{mcfee14librosa}, using 16 kHz sampling rate, 512-sample window size with no temporal overlaps.

\subsection{Model Implementation}

We fix the settings for the early and late convolution layers as follows. For the early convolution layers, we use one (64, 8, 4) convolution layer. For the late convolution layers, we use a stack of two (512, 1, 1) convolution layers, followed by a (50, 1, 1) convolution layer for tag prediction outputs.

When persistence landscapes are required, a $(C_0, C_1, P, Q)=(0, 5, 5, 10)$ persistence layer is the default setting for it. As there are 64 filters from the early convolution layer, the input to persistence will have 64 filters. Therefore, the default feature size is $5\times10\times64=3200$. In preliminary experiments, we found that the segment size $T$
%in the early convolution layer 
has an effect on the performance of PNN. %, as shown in Figure \ref{fig:segsize}. 
Slightly better result is achieved by setting the segment size to 64, but similar result can be obtained by choosing 32 or 128. In balance of performance and efficiency, we fix the segment size to 32.

In a PCNN, the middle convolution layer is a $(3200, 1, 32)$ convolution layer. Note that the max-pooling size is equal to the segment size in the persistence layer. Therefore, the outputs of the middle convolution layer and the persistence layer will have the same temporal size.

We implement our models with Theano and Lasagne\footnote{\url{https://lasagne.readthedocs.org/en/latest/}}. The CNNs are trained with back-propagation and AdaGrad \cite{duchi11adagrad}, 0.5 dropout rate, and 0.01 initial learning rate. A model is trained with 100 epochs. The parameters from the epoch that gives the best average per-class AUC on validation set is adopted. For reproducibility, the python source codes will be made publicly available at \url{http:\\removed for double-blinded review}.

\subsection{Results}
We compare the performance of CNN, PNN, and PCNN models with different settings. As the performance of a neural network model can be sensitive to initialization values, we execute the network training and evaluation 5 times for each setting and report the average results of the 5 trials. When assessing whether there is a significant performance difference between two models, we apply a one-tailed t-test under 0.1 confidence level on the results of the 5 trials. % of the two models.

\begin{table*}[htbp]
\caption{Evaluation of music auto-tagging on the MagnaTagATune dataset. Default: segment size=32; early conv length: 64. We use bold font to indicate the best result for the three types of models: CNN, PNN, and PCNN. ` Mid. conv. size' indicates the number of filters in the middle convolution layer. }
\begin{center}
\begin{tabular}{lcccccc}
\toprule
Model  & Mid. conv. size & $P$ (i.e.  & \multicolumn{ 2}{c}{Average AUC} & \multicolumn{ 2}{c}{MAP} \\
\cmidrule{4-7}
	&	& \# of $\lambda_k) $	& Per-class & Per-clip & Per-class & Per-clip \\
\midrule

CNN.3200		& 3200		& NA				& 0.8967		& \textbf{0.9333}	& \textbf{0.4187}	& 0.6808 \\
CNN.6400		& 6400		& NA				& \textbf{0.8972}	& \textbf{0.9333}	& 0.4173		& \textbf{0.6814} \\
\midrule
PNN ($P=1$)			& NA			& 1				& \textbf{0.8965} 	& 0.9315 		& \textbf{0.4147} 	& 0.6773 \\
PNN ($P=3$)		& NA			& 3				& 0.8959 		& \textbf{0.9322} 	& 0.4129 		& \textbf{0.6794} \\
PNN ($P=5$)		& NA			& 5				& 0.8949		& 0.9315		& 0.4121		& 0.6770 \\
PNN ($P=7$)		& NA			& 7				& 0.8931 		& 0.9300 		& 0.4082 		& 0.6744 \\
\midrule
PCNN ($P=1$) 		& 3200 		& 1				& 0.9010 		& \textbf{0.9366} 	& 0.4262 		& 0.6906 \\
PCNN ($P=3$) 		& 3200 		& 3				& 0.9006 		& 0.9364 		& 0.4264 		& \textbf{0.6915} \\
PCNN ($P=5$)		& 3200 		& 5				& \textbf{0.9013}	& 0.9365		& \textbf{0.4267}	& 0.6902 \\
PCNN ($P=7$)		& 3200 		& 7				& 0.9001 		& 0.9357 		& 0.4230 		& 0.6883 \\

\bottomrule
\end{tabular}
\end{center}
\label{tab:performance5}
\end{table*}

\begin{table*}[htbp]
\caption{Comparison with the result of existing work}
\begin{center}
\begin{tabular}{ccccccc}
\toprule
Number of tags & Model & \multicolumn{ 2}{c}{Average AUC} & \multicolumn{ 2}{c}{MAP} \\
\cmidrule{3-6}
&	& Per-class & Per-clip & Per-class & Per-clip \\
\midrule
50 	& Dieleman \emph{et al.} \cite{dieleman13multiscale}	& 0.898 		& NA 		& NA 		& NA \\
50	& Dieleman \emph{et al.} \cite{dieleman14icassp}	& 0.8815 		& NA 		& NA 		& NA \\
50	& PCNN ($P=1$) 							& 0.9010 		& 0.9366 	& 0.4262 	& 0.6906 \\
50	& PCNN ($P=5$) 							& \textbf{0.9013}	& 0.9365	& 0.4267	& 0.6902 \\
\midrule
160	& Nam \emph{et al.} \cite{nam15}			& 0.888		& 0.956	& NA		& NA \\
\midrule
160 & PCNN ($P=5$)							& 0.8931		& 0.9535	& 0.2214	& 0.6126 \\
188 & PCNN ($P=5$)							& 0.8864		& 0.9564	& 0.1986	& 0.6088 \\

\bottomrule
\end{tabular}
\end{center}
\label{tab:comparison}
\end{table*}

First we notice that PNNs do not perform better than CNNs, as shown in Table \ref{tab:performance5}. Interestingly, the performance is decreasing as the number of $\lambda_k$ increases. We will get back to this issue later.

With the combination of convolution features and persistence landscape, PCNNs outperform the CNNs significantly. PCNN with $P=5$ achieves the best performance with 0.9013 average per-class AUC and 0.4267 average per-class MAP.

In Table \ref{tab:comparison}, we compare the performance of the proposed model with two results from Dieleman and Schrauwen \cite{dieleman13multiscale, dieleman14icassp}. The proposed model outperforms these two prior arts. We also present the results of PCNN ($P=5$) trained with the top 160 tags and all 188 tags, and the performance from Nam \emph{et al.} \cite{nam15} with 160 tags in the earlier version of MagnaTagATune. We cannot compare the two models fairly but it gives us a reference about the performance for a larger number of tags.

The persistence layer naturally produces features of large dimension. PCNN ($P=5$) has 6400-D in the middle. One may wonder if the improvements is purely from the large dimension. This should not be the case. As we can see in Table \ref{tab:performance5}, the CNN using 6400 filters is not as good as PCNN models even though they have the same size in the middle. Furthermore, our analysis shows that the results of 200, 400, 800, 1600, 3200, and 6400 filters are not significantly different. %, as shown in Figure \ref{fig:cnntrend}.

\begin{figure*}[!t]
\centering
\includegraphics[width=\textwidth]{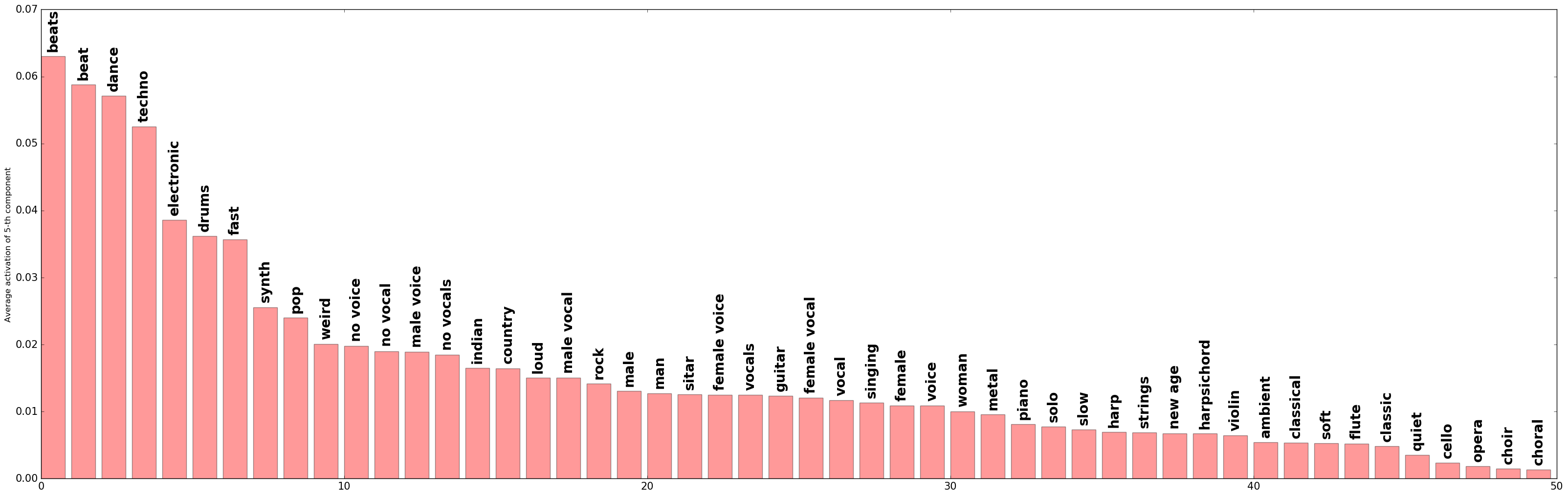}
\caption{Average value of the 5-th component of persistence landscapes}
\label{fig:5th}
\end{figure*}

An interesting problem is what kind of information is contained in persistence landscapes in the real data. By the definition of persistence landscape and observing simple signals such as those in Figure \ref{fig:sigpl}, we conjecture that the persistence landscape, as applied to music signals, may contain information about the number of strong beats or the number of onsets. To verify this, we compute 1) the average onset strength for each clip and 2) the average persistence landscape values over all sample values, all segments, and all 64 filters in a clip for each $\lambda_k$ and each clip from PCNN ($P=5$). The Pearson's correlation coefficients between the average onsets strength and the average persistence landscape are 0.7297, 0.9556, 0.9774, 0.9709, and 0.9560 for $\lambda_k$ with $k=1, 2, 3, 4, 5$ respectively. They are highly correlated, especially for larger $k$. In contrast, the correlation coefficient is 0.011 between the average onset strength and the average absolute values of the output of middle convolution layer of PCNN ($P=5$). We may see this property from another perspective. For a given tag, we compute the average persistence values for $\lambda_5$. By arranging them in descending order, as shown in Figure \ref{fig:5th}, we can see that those tags with more strong beats or more fast tempos are on the left.

Different pieces in persistence landscapes contribute to PNN and PCNN differently. To see the contribution of different pieces in persistence landscape to the late convolution layers, we compute the average parameter weights of the connections between persistence layer and the first layer of the late convolution layers in PNN ($P=5$) and PCNN ($P=5$), as shown in Table \ref{tab:average_weights}. We see that the middle pieces are more important for PNN, while the first piece has larger contributions than other pieces for PCNN. %and the contribution decreases as the $k$ increases. 
On the other hand, although the best performance is achieved by PCNN ($P=5$), we find the average AUCs from PCNN ($P=1$) and PCNN ($P=3$) are also quite high. These two observations raise a question about the usefulness of the later $\lambda_k$s.

\begin{table}[htbp]
\caption{Average parameter weights of PL to the first layer of the late convolution layers}
\begin{center}
\begin{tabular}{lccccc}
\toprule
 &  \multicolumn{ 5}{c}{$\lambda_k$} \\
\cmidrule{2-6}
 		& $k=1$	& $k=2$	& $k=3$	& $k=4$	& $k=5$ \\
\midrule
PNN ($P=5$) 		& 0.02752	& 0.02930	& 0.02985	& 0.02951	& 0.02767 \\
PCNN ($P=5$) 	& 0.02020 	& 0.01995	& 0.01940	& 0.01860	& 0.01736 \\
\bottomrule
\end{tabular}
\end{center}
\label{tab:average_weights}
\end{table}

We look into the performance tag-wisely. It turns out that the PCNNs incorporating later pieces perform consistently better in ``classical,'' ``slow,'' ``soft,'' and ``choir,'' the tags that characterize gentler music. In contrast, PCNN ($P=1$) consistently performs better in tags related to human voices and electronic music. Our conjecture is that the later pieces of persistence landscape might  signify the absence of the more fluctuated part of the signal, which help the classification of gentler music.

\section{Discussion and conclusions} \label{sec:conclusion}
In this paper, we have presented a CNN model that incorporates the topological tool  persistence landscape. We show empirically that the use of a dedicated persistence layer in the middle of a CNN model can perform similarly as a pure CNN model, and that the combination of the two models greatly improves the discriminative power. Evaluating on the MagnaTagATune dataset for music auto-tagging, the combined model outperforms the state-of-the-art models by a great margin.

As we observe in Section \ref{sec:exp}, different parts of the persistence landscape could help the classification in different ways. Using the same setting of persistence layer for all tags might lead to sub-optimal results. One way to utilize this property is to train multiple models and then combine the results.

In this paper, we only use the information of 0-homology classes which characterize connected components as we assume 2-connectivity. This provides an efficient implementation of persistence landscape. However, it also loses the information of higher order homology classes. An interesting future work is to re-formulate the used complex to account for higher order shape information.

% \fi
\IEEEpeerreviewmaketitle

\ifCLASSOPTIONcaptionsoff
  \newpage
\fi



%\bibliographystyle{IEEEtran}

\bibliographystyle{IEEEtran}
%\bibliography{IEEEabrv,JYLiuBib}
\bibliography{pnn}


%\end{IEEEbiography}


%\newpage


%\vfill


\end{document}